\documentclass[11pt]{article}

\usepackage[margin=1in]{geometry}
\usepackage[T1]{fontenc}
\usepackage{lmodern}
\usepackage{cite}
\usepackage{amsmath,amssymb,amsfonts}
\usepackage{graphicx}
\usepackage{textcomp}
\usepackage{booktabs}
\usepackage{array}
\usepackage{tabularx}
\usepackage{multirow}
\usepackage{url}
\usepackage{microtype}
\usepackage[hidelinks]{hyperref}

\title{FACET: Facet Aware Contrastive Embeddings for Traits --- Big Five Estimation from Life Narratives}

\author{
Balakrishnan Arumugam\textsuperscript{1,*} \quad
Priyen Parekh\textsuperscript{1,*} \quad
Joshua Oltmanns\textsuperscript{2} \quad
Mehak Gupta\textsuperscript{1}\\[2ex]
{\small \textsuperscript{1}Department of Computer Science, Southern Methodist University, Dallas, TX, USA}\\
{\small \texttt{barumugam@smu.edu}, \texttt{priyenkeyurp@smu.edu}, \texttt{mehakg@smu.edu}}\\[1ex]
{\small \textsuperscript{2}Department of Psychological and Brain Sciences,}\\
{\small Washington University in St.\ Louis, St.\ Louis, MO, USA}\\
{\small \texttt{j.oltmanns@wustl.edu}}\\[1ex]
{\small \textsuperscript{*}Contributed equally to this work.}\\
{\small Corresponding author: Mehak Gupta (\texttt{mehakg@smu.edu})}
}

\date{}

\begin{document}

\maketitle

\begin{abstract}
Personality is only weakly and inconsistently signaled by written language, and existing text-based methods typically combine language-model and psycholinguistic representations at the input or decision level. It remains open whether a more deliberate combination of the two views, and a decomposition of the prediction target, can recover more of the signal. We present a three-phase framework for estimating Big Five personality domain scores from life-narrative interview text. It encodes text and psycholinguistic features separately and combines them by concatenation or by a learned weighting, decomposes each domain into contrastively trained facet embeddings instead of predicting the domain directly, and pools the facet embeddings by attention into a domain-level prediction. We compare end-to-end with staged training and ablate the input modalities, their combination, and the facet-pooling strategy. On 1{,}332 participants from the St.\ Louis Personality and Aging Network (SPAN) study, facet embeddings reach AUC up to 0.740, and domain-level $R^2$ reaches 0.328, highest for Openness. End-to-end training gives lower error than staged training in every domain, and attention pooling gives the lowest error in every domain. Combining the two views lowers error relative to text alone for Conscientiousness and Agreeableness, whereas a learned weighting of the views shows no consistent advantage over concatenation. The framework provides facet-level predictions that a domain-only model cannot, and it improves modestly on a regression from the raw features in four of the five domains.
\end{abstract}


\section{Introduction}
Personality is a stable configuration of individual differences in behavior, cognition, and emotion, and understanding it has consequences well beyond psychological research itself. Personality assessment contributes to psychotherapy and clinical diagnosis~\cite{miller2015understanding}, career development and vocational guidance~\cite{hu2021mediating}, social science research~\cite{phan2021personality}, and personalized services such as targeted marketing and recommendation systems~\cite{shumanov2021making}. As personality assessment moves beyond the clinic and into everyday digital interfaces, the ability to estimate personality automatically, at scale, and without requiring a respondent to complete a structured instrument has become increasingly valuable.

The widespread availability of naturalistic text, from social-media posts to narrative interviews, has made language a natural substrate for automated assessment. Personality is known to surface, even if only weakly, in the words people choose~\cite{pennebaker1999styles}, and this observation has motivated two decades of work on predicting personality directly from text rather than from questionnaire responses. Early approaches relied on hand-crafted psycholinguistic features fed into classical machine learning models~\cite{mairesse2007linguistic}. More recent work has shifted toward deep learning, using document-level neural encoders to extract personality signal directly from raw text~\cite{lynn2020hierarchical,zhou2022attention}, and, increasingly, toward enriching these representations with explicit psychological knowledge in the form of lexicon features or knowledge graphs~\cite{yang2021psycholinguistic,zhu2022contrastive}.

Two kinds of representation underlie much of this work. Psycholinguistic lexicons such as LIWC~\cite{tausczik2010liwc} and the NRC emotion lexicon~\cite{mohammad2013nrc} reduce a document to counts over interpretable categories, such as pronoun use, cognitive-process words, and affect terms. They offer a transparent link between specific word choices and personality but discard word order and broader context. Pretrained language models~\cite{devlin2019bert} instead encode a passage into a dense representation that captures far more of this contextual structure, at the cost of the lexicon's interpretability. A natural strategy is to combine the two, and prior work has done so, typically with a single fixed rule applied identically to every input, either by concatenating the two feature sets or by combining their separate predictions at the decision level~\cite{mehta2020bottomup}. Whether such a fixed combination is the right assumption remains largely unexamined.

A second question concerns the prediction target. Each Big Five domain is composed of six facets, which describe more specific tendencies than the domain they belong to and are of interest in their own right for interpretation and assessment. It is unclear whether a domain should be predicted directly from a single combined representation, or whether first decomposing it into its facets yields a better representation and also provides facet-level predictions that a domain-only model cannot.

This paper addresses both questions with a three-phase framework for predicting continuous Big Five domain scores from life-narrative interview text. The first phase encodes the text and psycholinguistic views separately and combines them either by concatenation or by a learned weighting that is specific to each participant and each facet, so that the relative contribution of each view can vary from person to person. The second phase decomposes each Big Five domain into its six constituent facets and learns a separate embedding for each facet with a label-supervised contrastive objective, which draws participants who share a facet's high or low label together and pushes participants on opposite sides apart. The third phase combines the six facet embeddings of a domain by attention into a single domain-level representation, from which the continuous domain score is predicted. The attention lets the model learn whether facets contribute equally or unequally instead of fixing their weights in advance.

We train this architecture under two regimes. In joint training, the contrastive and regression losses are optimized together, so the regression loss also shapes Phases 1 and 2. In staged training, Phases 1 and 2 are optimized together under the contrastive objective and frozen before Phase 3 is trained. Comparing the two tests whether end-to-end gradient flow improves the learned representations.

This paper makes the following contributions.
\begin{itemize}
\item We propose a three-phase framework that decomposes personality prediction into facet-level contrastive learning followed by attention-based pooling, rather than regressing a domain score from an undifferentiated representation. This yields facet-level predictions that a domain-only model cannot provide.
\item We compare concatenation of the text and psycholinguistic views with a learned weighting of the two views, computed separately for each participant and each facet, and find that the benefit of the weighting depends on the trait and is not uniform.
\item We compare joint end-to-end training with staged training of the same architecture under identical optimization settings, and find that joint training, in which the regression loss reaches the facet and modality-combination representations directly, outperforms staged training across most domains.
\item We evaluate the framework on 1{,}332 participants from the St.\ Louis Personality and Aging Network (SPAN) study, reporting facet-level classification and domain-level regression, with ablations over the input modalities, their combination, the facet-pooling mechanism, and the removal of Phases 1 and 2.
\end{itemize}

On this dataset, the full model recovers Openness most reliably ($R^2=0.328$) and Neuroticism only marginally ($R^2=0.008$), consistent with the uneven and generally weak signal that text carries about personality. The remainder of this paper reviews related work, describes the framework and its training regimes, presents the experimental setup, and reports and discusses the results.

\section{Related Work}
\label{sec:related}

\subsection{Personality Estimation from Text}

Linguistic style behaves as a stable individual difference that covaries with personality~\cite{pennebaker1999styles}, and this observation underlies most automated approaches. Early systems summarized a document with closed-vocabulary lexicons such as LIWC~\cite{tausczik2010liwc} and passed the category frequencies to classical learners~\cite{mairesse2007linguistic}. Large-scale analyses of bloggers and social-media users extended this to individual words and open-vocabulary features~\cite{yarkoni2010personality,schwartz2013personality}. Neural models later replaced hand-built features with learned document representations, using convolutional and recurrent encoders~\cite{majumder2017deep,zhou2022attention}, hierarchical attention over a user's messages~\cite{lynn2020hierarchical}, and pretrained language models~\cite{devlin2019bert}. Most of this work relies on student essays or social-media content, and the prediction target varies widely. PsyAttention~\cite{zhang2024psyattention}, for instance, reports accuracy on binary high/low labels for the Essays and Myers-Briggs Type Indicator (MBTI) corpora, whereas HG-PerCon~\cite{li2024hgpercon} regresses continuous trait scores on YouTube, PAN, and MyPersonality data. Differences in corpus, label format, and metric make reported numbers difficult to compare across studies.

Life-narrative interviews are a different kind of text. Participants narrate their own lives in response to open-ended prompts, and in the SPAN study the transcripts average about 2,500 words~\cite{hussain2026language}. Language-based models have recently been applied to such interviews to estimate personality traits and pathology~\cite{oltmanns2026language}. Hussain et al.~\cite{hussain2026language} fine-tune a transformer encoder on overlapping 512-token windows, aggregate the window embeddings with an attention-equipped recurrent network, and regress each Big Five domain from the transcript, using the window attention to locate trait-relevant passages. We use the same interview source with a frozen language model and study a complementary question, namely how a psycholinguistic view should be combined with the text, and whether predicting each domain through its facets gives a better-structured representation.

\subsection{Combining Psycholinguistic and Learned Representations}

Lexicon features and language-model embeddings carry partly different information. The former are interpretable but ignore word order and context, whereas the latter capture context at the expense of transparency. Several studies therefore use both, for example by combining psycholinguistic features with BERT embeddings under standard classifiers~\cite{mehta2020bottomup}, or by pairing BERT with recurrent networks run over within-text contours of psycholinguistic features~\cite{kerz2022pushing}. Graph-based methods inject LIWC knowledge more structurally, through a tripartite graph with a BERT-based initializer~\cite{yang2021psycholinguistic} or through LIWC-based augmentations of a user's post graph~\cite{zhu2022contrastive}. PsyAttention~\cite{zhang2024psyattention} argues that hundreds of tool-derived features introduce redundancy and noise. It discards about 85\% of them by correlation analysis and encodes the remainder with a transformer that has no positional encoding, since the features have no natural order. FT-Transformer~\cite{gorishniy2021revisiting} adapts self-attention~\cite{vaswani2017attention} to tabular inputs in a comparable way by mapping each scalar feature to its own token, and our psycholinguistic encoder uses the same feature tokenizer. HG-PerCon~\cite{li2024hgpercon} instead aligns a semantic view built from a user's post history with a view derived from a psychological knowledge graph, using a cross-view contrastive objective. Replacing this alignment with plain concatenation raised its overall prediction error on each of its three datasets.

A related design question is how much each view should contribute. Concatenation applies the same rule to every input, whereas gated fusion lets the input decide. The gated multimodal unit learns multiplicative gates that set each modality's influence~\cite{arevalo2017gated}, and a recent clinical prediction model uses patient-specific routing weights because the relevance of each modality varies across patients and outcomes~\cite{hooman2026multimodal}. Within personality computing, PsyAttention rescales the text and psycholinguistic embeddings with weights computed from each embedding before concatenating them, and removing this weighting lowered accuracy on every Big Five trait in the Essays corpus~\cite{zhang2024psyattention}. That evidence comes from binary classification, and it remains to be established whether a participant-specific weighting improves on a fixed combination when the target is a continuous score, and whether any benefit holds across all five traits. Our first phase compares the two directly. For each facet head, a small MLP scores the two views from their concatenation, a softmax normalizes the two scores, and the weighted views are summed.

\subsection{Contrastive Supervision, Facets, and Training Regimes}

Contrastive objectives shape a representation by pulling designated positive pairs together and pushing negatives apart, either with a margin on pairwise distances~\cite{hadsell2006dimensionality} or with InfoNCE-style losses over many negatives~\cite{oord2018cpc,chen2020simclr}. CLIP~\cite{radford2021clip} uses the latter to align two modalities in a shared space. The central design choice is what counts as a positive pair. Supervised contrastive learning treats examples that share a class label as positives~\cite{khosla2020supcon}. In personality prediction, contrastive terms have mostly been self-supervised. HG-PerCon pairs the two views of the same user and treats other users in the batch as negatives~\cite{li2024hgpercon}, and the contrastive graph transformer network~\cite{zhu2022contrastive} contrasts LIWC- and semantics-based augmentations of a post graph. In both, trait labels enter only through the downstream prediction loss, so the contrastive term rewards agreement between views of one user and not similarity between users with similar trait levels. We instead let facet labels define the positives.

The five-factor model is also hierarchical. In the Revised NEO Personality Inventory (NEO-PI-R), each domain is measured through six narrower facets~\cite{costa1992neo}. Text-based systems nonetheless tend to predict domains directly, as binary labels~\cite{zhang2024psyattention} or continuous scores~\cite{li2024hgpercon}, including from life narratives~\cite{hussain2026language}. Facet recognition from text is less common. Dos Santos and Paraboni~\cite{dossantos2019facets} trained independent binary classifiers for ten facets inferred from a short Big Five inventory, using Brazilian Facebook posts, and treated each facet as an end target. To the best of our knowledge, using facet supervision as an intermediate representation from which domain scores are then predicted has not been examined in this literature.

Multi-stage models must also decide whether to train their stages separately or together. PsyAttention fine-tunes BERT toward the psycholinguistic feature vector in a separate step and freezes it before training the classifier~\cite{zhang2024psyattention}. HG-PerCon compares contrastive pretraining followed by fine-tuning with joint training under a weighted sum of the contrastive and regression losses, and reports the staged variant to be slightly better on all three of its datasets~\cite{li2024hgpercon}. Its contrastive objective is self-supervised, whereas ours is supervised by facet labels, so which regime suits label-supervised contrastive training remains to be examined.

\subsection{Positioning of This Work}

In the studies reviewed here, domains are predicted directly, and facets, when modeled at all, are separate targets. Learned weighting of two views has been evaluated mainly for classification, and the comparison of staged with joint training has been made for self-supervised contrastive objectives. We examine each of these on SPAN life narratives. Our framework combines the two views under fixed and learned rules, predicts each domain through six contrastively supervised facet embeddings, and compares joint with staged training of the same architecture.

\section{Methodology}

\subsection{Overview}

\begin{figure}[t]
\centering
\includegraphics[width=\textwidth]{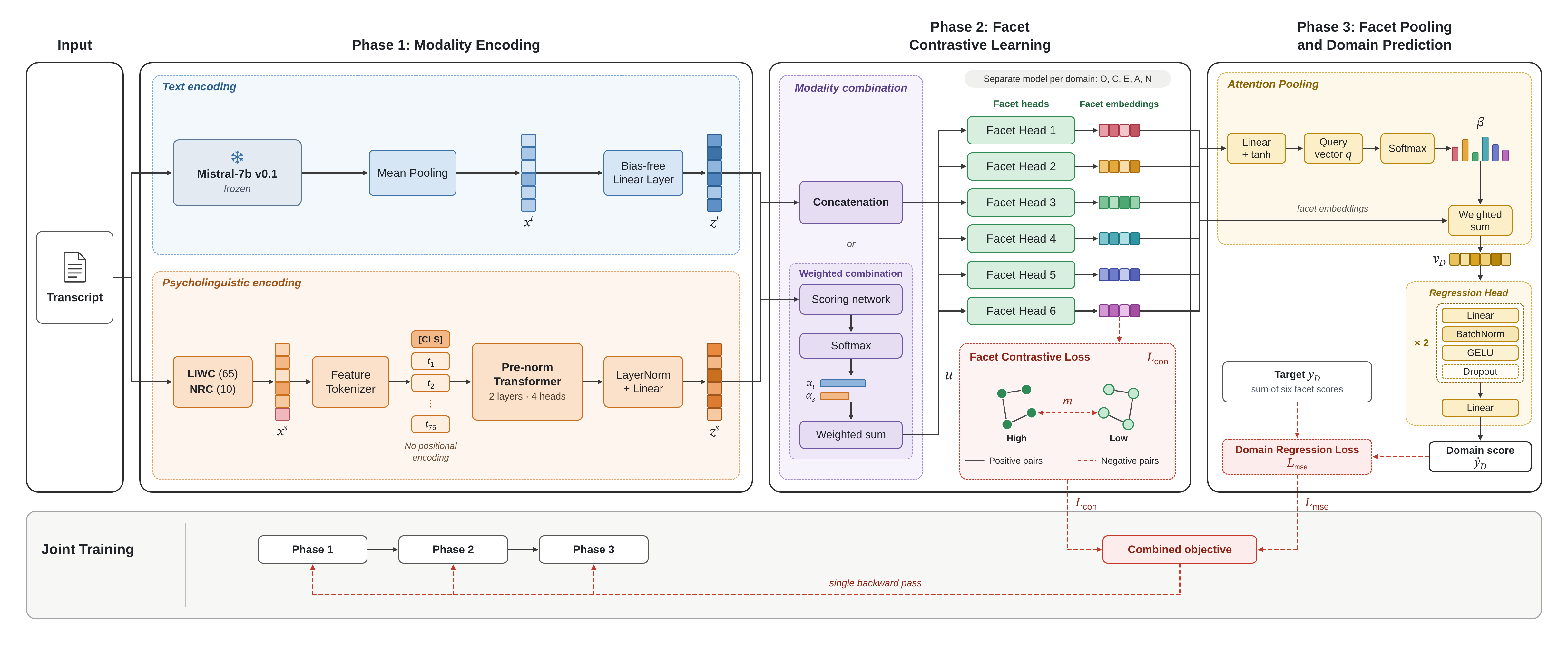}
\caption{\textbf{Phase 1 encodes the text and psycholinguistic views, Phase 2 combines them and learns six contrastive facet embeddings, and Phase 3 pools them with attention to predict the domain score.}}
\label{fig:facet_arch}
\end{figure}

We propose a three-phase framework for predicting continuous Big Five personality domain scores from text by integrating language-model embeddings with psycholinguistic features. In Phase 1, the textual and psycholinguistic views of each participant are encoded separately into a unit-normalized 256-dimensional space.
In Phase 2, the two embeddings are first combined, either by concatenation, which gives every facet head the same input, or by a learned, participant-specific weighted sum computed separately for each facet head. Six facet-specific embeddings are learned for each personality domain using a label-supervised, margin-based contrastive objective. The facet-level embeddings are trained to distinguish participants with high and low scores on the corresponding facets, thereby incorporating facet-level personality information into the learned representation. In Phase 3, the six facet embeddings are combined using an attention mechanism to form a domain-level representation, which is then used to predict the continuous domain score. Fig.~\ref{fig:facet_arch}  shows the full architecture.

We evaluate the framework under two training regimes. In the joint strategy, all three phases are optimized simultaneously using a combined contrastive and regression objective. In the staged strategy, Phases 1 and 2 are first trained together under the contrastive objective and frozen, and Phase 3 is then trained under the regression objective.

We further conduct ablation experiments to examine the contributions of the input views, modality-combination strategy, and facet-pooling mechanism to personality prediction.

\subsection{Problem Formulation}

Let $p \in \{1,\ldots,N\}$ denote a participant. Each participant is associated with a transcript from which we obtain a text embedding $x^{t}_{p} \in \mathbb{R}^{4096}$ and a psycholinguistic feature vector $x^{s}_{p} \in \mathbb{R}^{75}$. The psycholinguistic vector is constructed by concatenating LIWC and NRC lexicon features. Personality is represented using the Big Five domains $D \in \{O,C,E,A,N\}$, where each domain consists of six facets.

We denote the score on facet $i$ of domain $D$ by $f_{D,i}$, for $i=1,\ldots,6$, and the continuous domain score by $y_D$. Facet scores are converted into binary high/low labels using the median of the corresponding facet scores over the training participants. These labels provide supervision for the contrastive learning objective, encouraging participants with the same high/low label to have similar facet representations and participants with different labels to have dissimilar representations.

For each domain, the proposed framework learns six facet-specific representations per participant,
\begin{equation}
g_{D,i} \in \mathbb{R}^{256}, \qquad i=1,\ldots,6,
\end{equation}
where each representation corresponds to one facet of the domain. The six facet representations are subsequently aggregated into a single domain-level representation per participant,
\begin{equation}
v_D \in \mathbb{R}^{256},
\end{equation}
which is used to predict the continuous domain score $\hat{y}_D$.

A separate model is trained for each of the five personality domains, and the models do not share parameters. The overall learning objective combines the facet-level contrastive loss with the domain-level regression loss. The specific formulation of these losses and their optimization under the joint and staged training strategies are described in the subsequent sections.

\subsection{Phase 1: Multimodal Combination}\label{sec:phase1}

The goal of this phase is to encode the text and psycholinguistic views of a participant into embeddings of the same dimensionality, so that they can be combined for a downstream task. The two views are encoded separately, with different architectures suited to their structure, and are only brought together at the input of the Phase 2. This phase carries no objective of its own. The text projection $W_t$ and the psycholinguistic encoder are trained by gradients from Phase 2 and, under joint training, also from Phase 3.

\subsubsection{Text encoding}

Each transcript is passed through a frozen pretrained Mistral-7B-v0.1 model~\cite{jiang2023mistral} loaded with the Hugging Face transformers library, and the final-layer hidden states are mean pooled over tokens to give the text embedding $x^{t}_{p} \in \mathbb{R}^{4096}$. This embedding is projected to 256 dimensions by a single bias-free linear layer and normalized to unit length,
\begin{equation}
z^{t}_{p} = \frac{W_t\, x^{t}_{p}}{\lVert W_t\, x^{t}_{p} \rVert_2}, \qquad W_t \in \mathbb{R}^{256 \times 4096}.
\label{eq:textproj}
\end{equation}

\subsubsection{Psycholinguistic encoding}

The psycholinguistic view $x^{s}_{p} \in \mathbb{R}^{75}$ is formed by concatenating 65 LIWC category scores with 10 NRC emotion scores.

We encode this view with a transformer, which operates over a sequence of vectors rather than raw scalars. To construct this sequence, we use a feature tokenizer~\cite{gorishniy2021revisiting}, which maps each of the 75 features from a single scalar into its own $d$-dimensional token, using a weight and bias learned separately for that feature,
\begin{equation}
t_{j} = x^{s}_{p,j}\, w_{j} + b_{j}, \qquad w_{j}, b_{j} \in \mathbb{R}^{d}, \qquad j = 1, \dots, 75,
\label{eq:tokenise}
\end{equation}
with $d = 128$. A learnable [CLS] token $t_{\mathrm{[CLS]}} \in \mathbb{R}^{d}$ is prepended to these tokens, giving the transformer's input sequence $T^{(0)} = \big(t_{\mathrm{[CLS]}} \mid t_1 \mid \cdots \mid t_{75}\big) \in \mathbb{R}^{76 \times d}$. No positional encoding is added, since there is no temporal or sequential relationship between the features.

This sequence is passed through $L = 2$ pre-norm transformer layers with four attention heads, feed-forward inner width $2d$, GELU activation, and dropout $0.1$, allowing the encoder to model dependencies between features. The output at the [CLS] position, $h_{\mathrm{[CLS]}}$, is layer-normalized, projected to 256 dimensions and normalized to unit length to obtain the psycholinguistic embedding,
\begin{equation}
z^{s}_{p} = \ell_2\!\big(W_s\,\mathrm{LN}(h_{\mathrm{[CLS]}}) + b_s\big), \qquad W_s \in \mathbb{R}^{256 \times d},
\label{eq:psyproj}
\end{equation}
where LN denotes layer normalization and $\ell_2(\cdot)$ divides a vector by its Euclidean norm.

\subsubsection{Modality combination}

Once the text and psycholinguistic embeddings $z^{t}_{p}$ and $z^{s}_{p}$ are obtained, they are combined at the input of each Phase 2 facet head. We use one of two combination methods, compared in the ablation analysis.

The first is concatenation. The two unit-normalized embeddings are stacked into a single vector,
\begin{equation}
u_p = [\, z^{t}_{p} ; z^{s}_{p} \,] \in \mathbb{R}^{512},
\label{eq:concat}
\end{equation}
which is passed unchanged to every facet head of Phase 2.

The second is a learned, participant-specific weighted combination of the two embeddings, computed separately for each facet head. For facet $i \in \{1,\dots,6\}$, a scoring network with its own parameters reads both embeddings together and outputs one score per modality,
\begin{equation}
\begin{bmatrix} e^{(i)}_{t,p} \\ e^{(i)}_{s,p} \end{bmatrix} = W_2^{(i)} \tanh\!\big(W_1^{(i)} [\, z^{t}_{p} ; z^{s}_{p} \,] + b_1^{(i)}\big) + b_2^{(i)},
\label{eq:modscore}
\end{equation}
where $W_1^{(i)} \in \mathbb{R}^{128\times 512}$, $b_1^{(i)} \in \mathbb{R}^{128}$, $W_2^{(i)} \in \mathbb{R}^{2\times 128}$ and $b_2^{(i)} \in \mathbb{R}^2$ are learnable. The parameters $W_2^{(i)}$ and $b_2^{(i)}$ are initialized to zero, so both modalities start with equal weight. The scores are then converted into normalized weights by a softmax,
\begin{equation}
\alpha^{(i)}_{m,p} = \frac{\exp\big(e^{(i)}_{m,p}\big)}{\sum_{m' \in \{t,s\}} \exp\big(e^{(i)}_{m',p}\big)}, \qquad m \in \{t,s\},
\label{eq:modweight}
\end{equation}
where $\alpha^{(i)}_{t,p}$ and $\alpha^{(i)}_{s,p}$ are the learned weights given to the text and psycholinguistic embeddings and satisfy $\alpha^{(i)}_{t,p}+\alpha^{(i)}_{s,p}=1$. The input to facet head $i$ is their weighted sum, normalized to unit length,
\begin{equation}
u^{(i)}_p = \ell_2\!\big(\alpha^{(i)}_{t,p}\, z^{t}_{p} + \alpha^{(i)}_{s,p}\, z^{s}_{p}\big) \in \mathbb{R}^{256}.
\label{eq:modsum}
\end{equation}
Because each scoring network has its own parameters, the weights, and hence the input to each head, can differ across facets, and each scoring network is trained by the loss of its own facet.

In the single-modality ablations, $u_p$ is simply $z^{t}_{p}$ or $z^{s}_{p}$ (256-dimensional), and no combination is applied.

\subsection{Phase 2: Facet Contrastive Learning}
\label{sec:phase2}

The embeddings $z^{t}_{p}$ and $z^{s}_{p}$ obtained in Phase 1 capture information from the textual and psycholinguistic views. Because each personality domain is composed of six distinct facets, we first combine the two embeddings and then map them into six facet-specific representations.

For each domain $D$, we use one projection head for each of its six facets, and all six heads read the same embeddings $z^{t}_{p}$ and $z^{s}_{p}$. With concatenation, every head receives the same input $u_p \in \mathbb{R}^{512}$ \eqref{eq:concat}. With the weighted combination, head $i$ receives its own input $u^{(i)}_p \in \mathbb{R}^{256}$, computed by its own scoring network \eqref{eq:modscore}--\eqref{eq:modsum}. Each head is a two-layer projection network (hidden width 512, GELU activation, layer normalization, and dropout 0.2) that maps its input to a unit-normalized facet embedding $g_{D,i,p} \in \mathbb{R}^{256}$.

\textbf{Facet contrastive loss.} The six facet embeddings are learned using a supervised, margin-based contrastive objective. For each facet $i$ of domain $D$, a pair of participants $(a,b)$ is assigned the label $y_{a,b}=1$ if both fall on the same side of the facet median (both high or both low) and $y_{a,b}=0$ otherwise. Pairs with $y_{a,b}=1$ are positive pairs and pairs with $y_{a,b}=0$ are negative pairs. Pairs are sampled once from training participants only, with up to 2,000 high--high and 2,000 low--low positive pairs per facet and two low participants as negatives for each high participant. At each step, 32 pairs are drawn per facet from this pool, independently of the participant batch used for $\mathcal{L}_{\mathrm{mse}}$. The distance between the facet embeddings of a pair is the cosine distance,
\begin{equation}
d(a,b) = 1 - \cos(g_{D,i,a}, g_{D,i,b}) \in [0,2],
\end{equation}
which is small when the two embeddings point in similar directions and large when they point in dissimilar directions.

For a batch $P$ of 32 pairs drawn for facet $i$, the contrastive loss follows the margin-based formulation of Hadsell et al.~\cite{hadsell2006dimensionality}:
\begin{multline}\label{eq:contrastive}
\mathcal{L}_i = \frac{1}{|P|}\sum_{(a,b)\in P}\Big[\,
y_{a,b}\, d(a,b)^2 \\
+ (1-y_{a,b})\max\big(0,\, m - d(a,b)\big)^2 \Big],
\end{multline}
where $m = 0.5$ is a fixed margin. For a positive pair, the loss is $d(a,b)^2$, which encourages the two embeddings to move closer together. For a negative pair, the loss is $\max(0,m-d(a,b))^2$, which penalizes the pair only while its distance is below the margin and therefore encourages sufficient separation between the embeddings. Because the pool contains roughly three to four positive pairs per negative pair, positive pairs make up most of each batch.

The overall contrastive loss for domain $D$ is the mean of the losses across its six facets:
\begin{equation}
\mathcal{L}_{\mathrm{con}} = \frac{1}{6}\sum_{i=1}^{6}\mathcal{L}_{i}.
\end{equation}

Although the six losses are combined into a single objective, each facet head, together with its scoring network when the weighted combination is used, receives contrastive signal only from its own loss $\mathcal{L}_i$, because the heads have independent parameters. Under joint training (Section~\ref{sec:joint-training}), every head also receives the gradient of $\mathcal{L}_{\mathrm{mse}}$ through the attention pooling. In contrast, the two shared Phase 1 components, the text projection $W_t$ and the psycholinguistic encoder, receive the combined learning signal from all six facet losses. Thus, the facet-specific heads learn their respective facet distinctions, while the shared encoders are shaped by information from all six facets.

\subsection{Phase 3: Facet Pooling and Domain Prediction}
\label{sec:phase3}

The six facet embeddings learned in Phase 2, $g_{D,1,p}, \dots,\allowbreak g_{D,6,p}$, are combined into a single representation before the domain score is regressed. For notational simplicity, the participant index $p$ is omitted from subsequent expressions even though quantities remain participant-specific. Because the six facets of a domain may not be equally recoverable from language, we use learned attention weights to pool the facet embeddings rather than assigning them equal weights.

\subsubsection{Attention pooling}

The domain representation is computed as a weighted sum of the six facet embeddings,
\begin{equation}
v_D = \sum_{i=1}^{6} \beta_{D,i}\, g_{D,i},
\label{eq:attnpool}
\end{equation}
where the attention weights are computed as
\begin{equation}
\beta_{D,i} = \frac{\exp\!\big(q^{\top}\tanh(U g_{D,i} + b_u)\big)}{\sum_{j=1}^{6}\exp\!\big(q^{\top}\tanh(U g_{D,j} + b_u)\big)}.
\label{eq:attnweight}
\end{equation}
This is the additive attention form used in hierarchical attention networks~\cite{yang2016hierarchical}. Here, $U \in \mathbb{R}^{128 \times 256}$, $q \in \mathbb{R}^{128}$, and $b_u \in \mathbb{R}^{128}$ are learned jointly with the rest of Phase 3. The query vector $q$ is initialized to zero, so that all six attention weights are equal at the start of training, $\beta_{D,i}=1/6$. Pooling therefore starts as an unweighted mean of the six facet embeddings, and subsequent training allows the model to learn a weight for each facet and each participant.

\subsubsection{Mean and concatenation pooling}
In the facet-pooling ablation, attention is replaced by two alternatives. Mean pooling sets $v_D=\frac{1}{6}\sum_{i=1}^{6}g_{D,i}$. Concatenation stacks the six embeddings and maps them to 256 dimensions with a two-layer MLP (hidden width 256, GELU), $v_D=\mathrm{MLP}(g_{D,1}\mid\cdots\mid g_{D,6})$. In both cases $v_D \in \mathbb{R}^{256}$ is passed to the same regression head as in attention pooling.

\subsubsection{Regression Head}
The pooled representation $v_D \in \mathbb{R}^{256}$ is passed through two blocks, each a linear layer followed by batch normalization, GELU activation and dropout (0.3), with output widths 256 and 128, and a final linear layer maps the 128-dimensional output to a scalar domain-score prediction.
\begin{equation}
\hat{y}_D = \mathrm{Regressor}(v_D).
\label{eq:regressor}
\end{equation}

\subsubsection{Domain Regression Loss}

The domain regression loss provides the training signal for the pooled representation, the attention weights, and the regression head. The target $y_{D,p}$ is the sum of participant $p$'s six facet scores for domain $D$, standardized with the mean and standard deviation of the training participants. For a participant batch $B$ of 32 participants, the loss is the mean squared error between the predicted and true standardized domain scores,
\begin{equation}
\mathcal{L}_{\mathrm{mse}} = \frac{1}{|B|}\sum_{p\in B}\big(\hat{y}_{D,p} - y_{D,p}\big)^2.
\label{eq:mse}
\end{equation}
Here the participant index $p$ is retained because the loss is computed over individual participants in the batch.

\section{Training Regimes}
\label{sec:training-regimes}
We use the same three-phase architecture but train it under two different training regimes. The regimes differ in how the facet contrastive loss $\mathcal{L}_{\mathrm{con}}$ and the domain regression loss $\mathcal{L}_{\mathrm{mse}}$ are combined and in which parameters each loss is allowed to update. Both regimes use the same participant-level split into training, validation and test sets (Section~\ref{sec:training-procedure}), with no participant appearing in more than one set. Thus, the two regimes are trained and evaluated under the same data partitioning.

\subsection{Joint Training}
\label{sec:joint-training}
In the joint regime, all three phases are optimized together under a single combined objective and a single backward pass per step:
\begin{equation}
\mathcal{L} = \lambda\, \mathcal{L}_{\mathrm{con}} + (1 - \lambda)\, \mathcal{L}_{\mathrm{mse}}.
\label{eq:joint}
\end{equation}
Here $\lambda$ weighs the contribution of each loss, and we use $\lambda=0.5$. At each step, a batch of 32 participants is used for $\mathcal{L}_{\mathrm{mse}}$ and 32 pairs per facet are used for $\mathcal{L}_{\mathrm{con}}$. A single optimizer updates the parameters of all three phases, using the settings in Section~\ref{sec:training-procedure}.

\subsection{Staged Training}
\label{sec:staged-training}
In the staged regime, each loss supervises only the phases it was designed for. In the first stage, Phases 1 and 2 are trained together under $\mathcal{L}_{\mathrm{con}}$ and then frozen. In the second stage, Phase 3 is trained under $\mathcal{L}_{\mathrm{mse}}$ alone on the facet embeddings produced by the frozen Phases 1 and 2. Optimization settings are those of Section~\ref{sec:training-procedure}.

\section{Experimental Setup}
\label{sec:setup}

\subsection{Data Preparation}
\label{sec:data}

\subsubsection{Source}
We use life-narrative interviews from the St.\ Louis Personality and Aging Network (SPAN) study~\cite{oltmanns2020personality}, in which older adults aged 55--64 from the St.\ Louis area recount their lives, describing key events and influential people. Personality was measured with the self-report NEO Personality Inventory--Revised (NEO-PI-R)~\cite{costa2008revised}, whose strong reliability and validity in this sample have been documented~\cite{oltmanns2020personality, wright2022prospective}. Each Big Five domain comprises six facets of eight items, rated from 0 to 4, so facet scores range from 0 to 32 and domain scores from 0 to 192. Of the 1{,}630 SPAN participants, we retain the $N=1{,}332$ for whom a transcript embedding, LIWC and NRC features, and complete NEO-PI-R facet scores are all available.

\subsubsection{Feature Construction}
Each participant is represented by two views. The text view is a frozen language-model embedding of the transcript (Section~\ref{sec:phase1}). The psycholinguistic view combines two dictionary-based feature sets that give interpretable counts of word categories.

\textbf{LIWC.} Linguistic Inquiry and Word Count (LIWC)~\cite{tausczik2010liwc} is a dictionary-based text analysis tool. It matches each word of a text against a validated dictionary and reports the percentage of words falling into each psychological and linguistic category, such as pronouns and other function words, and affective, cognitive, and social processes. LIWC scores were supplied with the SPAN data, so we did not compute them. They comprise one word-count column and percentage columns for the remaining categories, 65 columns in total.

\textbf{NRC.} The NRC Word--Emotion Association Lexicon~\cite{mohammad2013nrc} lists roughly 14{,}000 English words, each annotated by crowdsourcing for association with eight basic emotions (anger, anticipation, disgust, fear, joy, sadness, surprise, and trust, following Plutchik~\cite{plutchik1980general}) and two sentiments (negative and positive). NRC features were not part of the SPAN release. We computed them by matching the tokens of each transcript against the lexicon and dividing each category count by the transcript's word count, giving 10 features per transcript, which we joined to the LIWC features by participant identifier.

Concatenating the two sets gives the 75-dimensional vector $x^{s}_{p}$. All 75 columns are retained, with no feature selection or dimensionality reduction. Each of the 75 features, like each dimension of the text embedding $x^{t}_{p}$, is z-scored using the mean and standard
deviation of the training participants.

\subsubsection{Targets and Labels}
The regression target for each domain is the sum of its six facet scores. For contrastive learning, each facet score is also converted to a binary high/low label by a median split, with the median computed over the training participants. Participants at or above the median are labeled high, and those below it are labeled low.

\subsection{Training Procedure}\label{sec:training-procedure}
Participants are split once at the participant level into training (80\%) and test (20\%) sets, with 5\% of the training participants held out as a validation set for early stopping and checkpoint selection. The split is stratified by the high/low label of the domain's first facet, and the same split is used for all phases and configurations. Contrastive pairs are drawn from training participants only.

All parameters are optimized by AdamW at a learning rate of $10^{-3}$ with weight decay 0.01, a participant batch size of 32, cosine annealing over 100 epochs, and gradient-norm clipping at 1.0, in both training regimes and in every ablation. Training halts when the validation MSE does not improve for 20 consecutive epochs, and the checkpoint with the lowest validation MSE is restored before evaluation. Runs use a fixed seed ($42$) and deterministic cuDNN settings. Table~\ref{tab:hyper} lists the settings.

\begin{table}[!htbp]
\centering
\caption{Hyperparameters.}
\label{tab:hyper}
\small
\setlength{\tabcolsep}{4pt}
\begin{tabularx}{\columnwidth}{@{}l>{\raggedright\arraybackslash}X>{\centering\arraybackslash}X@{}}
\toprule
\textbf{Component} & \textbf{Parameter} & \textbf{Value} \\
\midrule
Embeddings & Dimension of $z^{t}_{p}$, $z^{s}_{p}$ & 256 \\
\midrule
Psy encoder & Token width $d$            & 128 \\
            & Attention heads            & 4 \\
            & Layers $L$                 & 2 \\
            & Feed-forward width         & 256 \\
            & Dropout                    & 0.1 \\
            & [CLS] token                & yes \\
\midrule
Modality weighting & Scorer hidden width & 128 \\
                   & Final-layer initialization & zero ($\alpha=0.5$) \\
\midrule
Facet head & Hidden width                & 512 \\
           & Output dimension            & 256 \\
           & Dropout                     & 0.2 \\
           & Margin $m$                  & 0.5 \\
           & Positive pairs per class (up to) & 2{,}000 \\
           & Negatives per high-label participant & 2 \\
           & Pairs per facet per step    & 32 \\
\midrule
Pooling & Attention scorer width         & 128 \\
        & Concatenation-MLP hidden width & 256 \\
\midrule
Regressor & Hidden widths                & 256, 128 \\
          & Dropout                      & 0.3 \\
\midrule
Training & Optimizer                     & AdamW \\
         & Learning rate                 & $10^{-3}$ \\
         & Weight decay                  & 0.01 \\
         & Loss weight $\lambda$         & 0.5 \\
         & Schedule                      & cosine, $T_{\max}=100$ \\
         & Batch size (participants)     & 32 \\
         & Maximum epochs                & 100 \\
         & Gradient clipping             & 1.0 \\
         & Early-stopping patience       & 20 (validation MSE) \\
         & Seed                          & 42 \\
\bottomrule
\end{tabularx}
\end{table}

\subsection{Types of Experiments}
\label{sec:experiments}

We evaluate our framework in three parts. First, we compare its prediction accuracy against existing text-based personality prediction methods. Second, we conduct ablation studies to isolate the contributions of individual design choices, including the modality inputs and their combination and the facet pooling strategy. Third, we evaluate the different training regimes and conduct phase-level ablations to examine the contribution of Phases 1--2 together. Unless stated otherwise, each ablation varies one factor while keeping the rest of the pipeline fixed at its default setting.

\subsubsection{Baseline Comparison}
\label{sec:baseline-comparison}
We compare the full model with four baselines trained on the same split and evaluated with the same metrics: (i) support vector regression on the 75 LIWC and NRC features~\cite{pennebaker2001liwc}; (ii) an MLP on BERT embeddings~\cite{devlin2019bert}, obtained by sliding a 512-token window over each transcript and mean-pooling the window embeddings; (iii) an MLP on the frozen Mistral embedding; and (iv) PsyAttention~\cite{zhang2024psyattention}, adapted to our regression task. The raw-feature MLP, which uses both views without the encoders, facet heads, or pooling, is reported in the ablation study as the model with Phases 1--2 removed.

\subsubsection{Design Choice Ablations}
\paragraph{Modality Ablation}
Our model integrates two data modalities, text and psycholinguistic features. To analyze the contribution of each, we test three input configurations: text only, psycholinguistic features only, and both modalities combined. Comparing the single-modality configurations against the combined configuration shows what each modality contributes on its own and what is gained by combining them. In the single-modality conditions, $u_p=z^{t}_{p}$ or $z^{s}_{p}$.

When both modalities are used, they can be combined in two ways.
The first is concatenation, in which the two embeddings are stacked into a 512-dimensional vector and passed directly to each facet head.
The second is a learned weighted combination (with a separate scorer per facet head), in which each modality is assigned a weight before the two are summed. Comparing these two combination methods shows whether letting each modality contribute unequally to the shared representation improves on a fixed combination rule.

\paragraph{Facet Pooling Ablation}
The six facet embeddings learned in Phase 2 can be pooled into a single domain-level representation in three ways: the unweighted mean, concatenation followed by a two-layer MLP, or learned attention, as described in Section~\ref{sec:phase3}. Comparing these three pooling strategies shows whether letting facets contribute unequally to the domain score, as attention pooling allows, improves on a fixed pooling rule.

\subsubsection{Training Regime and Phase Removal Ablation}
\label{sec:regime-phase-ablation}
This ablation examines two questions. The first is whether the architecture should be trained jointly or in stages (Sections~\ref{sec:joint-training} and \ref{sec:staged-training}). The second is whether the encoders, facet decomposition, and attention pooling contribute anything beyond a single regression model trained on the raw features. To answer the second question, we remove Phases 1 and 2 and train a regression MLP directly on the concatenated z-scored text and psycholinguistic features $[x^{t}_{p}; x^{s}_{p}] \in \mathbb{R}^{4171}$ (4{,}096 text dimensions plus 75 psycholinguistic features), with the same hidden widths as the Phase 3 head (256 and 128, with batch normalization, GELU and dropout 0.3), under $\mathcal{L}_{\mathrm{mse}}$ alone.

This gives three settings:
\begin{itemize}
\item \textbf{Joint.} All three phases are trained together under the combined objective.
\item \textbf{Staged.} Phases 1 and 2 are trained under $\mathcal{L}_{\mathrm{con}}$ and frozen, and Phase 3 is then trained under $\mathcal{L}_{\mathrm{mse}}$.
\item \textbf{Phases 1--2 removed.} The raw-feature MLP is trained under $\mathcal{L}_{\mathrm{mse}}$. With no phases to freeze and no contrastive loss, there is a single version of this setting.
\end{itemize}
Joint and staged training use the same architecture, data split, and optimization settings, so comparing them isolates whether the regression objective should be allowed to shape the earlier phases. Comparing each with the raw-feature MLP shows what the full pipeline adds.

\subsection{Evaluation Metrics}
Domain-level predictions are mapped back to the original 0--192 scale and compared with the test participants' scores using $R^2 = 1 - \sum_p (y_p - \hat y_p)^2 / \sum_p (y_p - \bar y)^2$ (which is negative when predictions are worse than the test mean), mean squared error (MSE), root mean squared error (RMSE), mean absolute error (MAE), and Pearson correlation $r$. Facet-level quality is measured with AUC of the prototype score $\sigma$, the difference in cosine similarity between a held-out participant's facet embedding and the high- and low-class prototypes computed from training embeddings, and with macro-averaged precision, recall, and F1 at the threshold $\sigma=0$.

\section{Results}

\subsection{Domain-Level Performance}

Table~\ref{tab:domains} summarizes the performance of the full model across the five Big Five domains on held-out participants.
The model shows the strongest association with the ground-truth domain scores for O, with an $R^2$ of 0.328 and Pearson correlation of 0.596.
C and E show moderate predictive performance, with $R^2$ values of 0.180 and 0.154, respectively.
A and N show lower explained variance, with $R^2$ values of 0.065 and 0.008; for N the explained variance is close to zero, and the Pearson correlation is only 0.148.
The facet-level classification results follow a similar pattern, with the highest mean AUC observed for O (0.694) and lower but above-chance AUC values for the remaining domains.
Overall, the results indicate that the framework captures personality-related information to different degrees across the five domains, with the strongest predictive signal observed for O and more limited signal for N.

\begin{table}[!htbp]
\centering
\caption{\textbf{Domain-level performance of the full model (text and psycholinguistic
embeddings combined by concatenation, attention-based facet pooling, joint
training) on the 266 held-out participants. AUC$_\mu$ is the mean facet-level
AUC across the domain's six facets; $R^2$, Pearson $r$, RMSE, and MAE are
computed on the original domain-score scale (0--192).}}
\label{tab:domains}
\small
\setlength{\tabcolsep}{4pt}
\begin{tabular}{@{}lccccc@{}}
\toprule
\textbf{Domain} & \textbf{AUC$_\mu$} & \textbf{{$R^2$}} & \textbf{{$r$}} &
\textbf{RMSE} & \textbf{MAE} \\
\midrule
O & 0.694 & 0.328 & 0.596 & 14.76 & 11.70 \\
C & 0.664 & 0.180 & 0.430 & 17.01 & 12.93 \\
E & 0.660 & 0.154 & 0.402 & 16.83 & 13.16 \\
A & 0.610& 0.065& 0.269& 13.21& 10.27 \\
N & 0.605 & 0.008 & 0.148 & 19.67 & 16.06 \\
\bottomrule
\end{tabular}
\end{table}

\subsection{Facet-Level Performance}
\label{sec:facet-quality-results}

Table~\ref{tab:facets} reports AUC, F1, precision, and recall for all thirty facets for the full model (concatenation, attention pooling, joint training), with classification metrics macro-averaged across the high and low classes.
AUC ranges from $0.557$ (A5) to $0.740$ (O6). With 266 test participants, the standard error of an AUC near 0.5 is about 0.04 , so the weakest facets (A5, A3, E5, N4, and N5, all below 0.58) lie within roughly two standard errors of chance, whereas all Openness and Conscientiousness facets lie more than three standard errors above it.
At the domain level, Openness has the highest mean facet AUC ($0.694$), followed by Conscientiousness ($0.664$) and Extraversion ($0.660$), while Neuroticism ($0.605$) and Agreeableness ($0.610$) have lower mean AUCs.
F1, precision, and recall are generally consistent within each facet, with no large differences among these measures.
Overall, the results indicate that the learned facet embeddings contain discriminative information for distinguishing high and low facet scores, although the strength of this signal varies across domains and facets.

\begin{table}[!htbp]
\caption{\textbf{Per-facet classification performance of the full model (text and
psycholinguistic embeddings combined by concatenation, attention-based facet
pooling, joint training) on the 266 held-out participants. Each held-out
participant is scored by the difference in cosine similarity to the high and
low class prototypes computed from training embeddings. AUC, F1, precision,
and recall are macro-averaged across the high and low classes.}}
\label{tab:facets}
\centering
\small
\begin{tabular}{@{}lcccc@{}}
\toprule
Facet & AUC & F1 & Prec & Rec \\
\midrule
\multicolumn{5}{@{}l}{\textit{Openness}}\\
O1 & 0.681& 0.625& 0.626& 0.626\\
O2 & 0.723& 0.670& 0.670& 0.675\\
O3 & 0.642& 0.615& 0.616& 0.620\\
O4 & 0.656& 0.594& 0.595& 0.598\\
O5 & 0.717&  0.683&  0.684& 0.686\\
O6 & 0.740& 0.656& 0.656& 0.665\\
\midrule
\multicolumn{5}{@{}l}{\textit{Conscientiousness}}\\
C1 & 0.732& 0.653& 0.653& 0.653\\
C2 & 0.646& 0.608& 0.608&  0.609\\
C3 & 0.653& 0.595& 0.596& 0.598\\
C4 & 0.658& 0.588& 0.587& 0.588\\
C5 & 0.661& 0.609& 0.609& 0.610\\
C6 & 0.631& 0.588& 0.588& 0.589\\
\midrule
\multicolumn{5}{@{}l}{\textit{Extraversion}}\\
E1 & 0.658& 0.591& 0.591& 0.592\\
E2 & 0.630& 0.605& 0.606& 0.605\\
E3 & 0.686& 0.640& 0.639& 0.640\\
E4 & 0.695& 0.651& 0.650& 0.655\\
E5 & 0.574& 0.539& 0.542& 0.544\\
E6 & 0.714& 0.646& 0.649& 0.646\\
\midrule
\multicolumn{5}{@{}l}{\textit{Agreeableness}}\\
A1 & 0.714& 0.677& 0.677& 0.677\\
A2 & 0.583& 0.570& 0.572& 0.572\\
A3 & 0.562& 0.533& 0.538& 0.541\\
A4 & 0.644& 0.606& 0.606& 0.609\\
A5 & 0.557& 0.543& 0.543& 0.544\\
A6 & 0.598& 0.575& 0.575& 0.575\\
\midrule
\multicolumn{5}{@{}l}{\textit{Neuroticism}}\\
N1 & 0.637& 0.605& 0.611&  0.614\\
N2 & 0.603& 0.561& 0.569& 0.571\\
N3 & 0.619& 0.593& 0.598& 0.600\\
N4 & 0.573& 0.555& 0.555& 0.555\\
N5 & 0.575& 0.527& 0.535& 0.534\\
N6 & 0.618& 0.609& 0.611& 0.611\\
\bottomrule
\end{tabular}
\end{table}

\begin{table}[!htbp]
\centering
\caption{Comparison with baselines on the 266 test participants. $R^2$ and MSE are computed on the original 0--192 domain scale. Best value per column in bold.}
\label{tab:baselines}
\setlength{\tabcolsep}{4pt}
\resizebox{\textwidth}{!}{%
\begin{tabular}{lcccccccccc}
\toprule
& \multicolumn{2}{c}{Openness} & \multicolumn{2}{c}{Conscientiousness} & \multicolumn{2}{c}{Extraversion} & \multicolumn{2}{c}{Agreeableness} & \multicolumn{2}{c}{Neuroticism} \\
\cmidrule(lr){2-3}\cmidrule(lr){4-5}\cmidrule(lr){6-7}\cmidrule(lr){8-9}\cmidrule(lr){10-11}
Model & $R^2\uparrow$ & MSE$\downarrow$ & $R^2\uparrow$ & MSE$\downarrow$ & $R^2\uparrow$ & MSE$\downarrow$ & $R^2\uparrow$ & MSE$\downarrow$ & $R^2\uparrow$ & MSE$\downarrow$ \\
\midrule
SVR on LIWC + NRC     & 0.134 & 280.66 & 0.097 & 318.58 & \textbf{0.156} & 282.49 & 0.037 & 179.84 & \textbf{0.027} & \textbf{379.31} \\
MLP on BERT           & 0.218 & 253.62 & 0.005 & 351.08 & \textbf{0.156} & \textbf{282.41} & $-0.342$ & 250.45 & $-0.069$ & 416.80 \\
MLP on Mistral        & 0.186 & 263.76 & 0.128 & 307.54 & 0.039 & 321.78 & $-0.222$ & 228.11 & $-0.135$ & 442.43 \\
PsyAttention          & $-0.030$ & 333.77 & $-0.067$ & 376.29 & $-0.007$ & 336.96 & $-0.015$ & 189.41 & $-0.055$ & 411.41 \\
\midrule
FACET (ours)          & \textbf{0.328} & \textbf{217.90} & \textbf{0.180} & \textbf{289.32} & 0.154 & 283.31 & \textbf{0.065} & \textbf{174.53} & 0.008 & 386.86 \\
\bottomrule
\end{tabular}%
}
\end{table}

\begin{table}[!htbp]
\centering
\caption{\textbf{Ablation results for domain-level personality prediction across the Big Five traits. Results examine the contribution of modality inputs, modality combination, facet pooling, and training regime with Phases 1--2 removed. For each trait, both $R^2$ and MSE are reported. Best value per column within each block in bold.}}
\label{tab:ablations}
\resizebox{\textwidth}{!}{%
\begin{tabular}{@{}lcccccccccc@{}}
\toprule
& \multicolumn{2}{c}{\textbf{O}} & \multicolumn{2}{c}{\textbf{C}} & \multicolumn{2}{c}{\textbf{E}} & \multicolumn{2}{c}{\textbf{A}} & \multicolumn{2}{c}{\textbf{N}} \\
\cmidrule(lr){2-3} \cmidrule(lr){4-5} \cmidrule(lr){6-7} \cmidrule(lr){8-9} \cmidrule(lr){10-11}
 & $R^{2}$ & MSE & $R^{2}$ & MSE & $R^{2}$ & MSE & $R^{2}$ & MSE & $R^{2}$ & MSE \\
\midrule
\multicolumn{11}{@{}l}{\textbf{Modality}}\\
Text only & 0.319&220.66&0.109&314.27&\textbf{0.178}&\textbf{275.17}&0.005&185.78&\textbf{0.010}&\textbf{386.23}\\
Psy only  & 0.129&282.44&0.048&336.00&0.101&301.10&0.014&184.09&0.005&388.11\\
Both (concatenation) & \textbf{0.328}&\textbf{217.90}&\textbf{0.180}&\textbf{289.32}&0.154&283.31&\textbf{0.065}&\textbf{174.53}&0.008&386.86\\
\midrule
\multicolumn{11}{@{}l}{\textbf{Modality Combination}}\\
Concatenation         & \textbf{0.328}&\textbf{217.90}&\textbf{0.180}&\textbf{289.32}&0.154&283.31&\textbf{0.065}&\textbf{174.53}&0.008&386.86\\
Weighted combination  & 0.241&246.22&0.122&309.69&\textbf{0.183}&\textbf{273.53}&0.049&177.46&\textbf{0.018}&\textbf{382.83}\\
\midrule
\multicolumn{11}{@{}l}{\textbf{Facet Pooling}}\\
Mean          &0.314&222.50 &0.128 &307.62 &0.133&290.20 &0.050&177.35 &0.003&388.76 \\
Concatenation &0.311&223.44 &0.107 &314.93 &0.132&290.60 &0.040&179.19 &0.003&388.81 \\
Attention     &\textbf{0.328}&\textbf{217.90} &\textbf{0.180}&\textbf{289.32} &\textbf{0.154}&\textbf{283.31} &\textbf{0.065}&\textbf{174.53} &\textbf{0.008}&\textbf{386.86}\\
\midrule
\multicolumn{11}{@{}l}{\textbf{Training Regime \& Phase Removal}}\\
Joint (full model)                    & \textbf{0.328}&\textbf{217.90}&\textbf{0.180}&\textbf{289.32}&\textbf{0.154}&\textbf{283.31}&\textbf{0.065}&\textbf{174.53}&0.008&386.86\\
Staged                                & 0.189&263.06&0.030&342.26&0.063&313.82&0.032&180.62&$-0.012$&394.63\\
Phases 1--2 removed (raw-feature MLP) & 0.250&243.07&0.121&310.19&0.093&303.78&0.014&184.07&\textbf{0.012}&\textbf{385.12}\\
\bottomrule
\end{tabular}%
}
\end{table}

\subsection{Baseline Comparison}
\label{sec:baseline-results}
Table~\ref{tab:baselines} compares the full model with four baselines trained and evaluated on the same split. FACET has the lowest MSE for Openness, Conscientiousness, and Agreeableness, and its $R^2$ is also the highest in these three domains. Relative to the best baseline in each domain, it reduces MSE by 14.1\% for Openness, 5.9\% for Conscientiousness, and 2.7\% for Agreeableness. For Extraversion, FACET is level with the best baselines, with an MSE 0.3\% above the lowest. For Neuroticism, all models explain almost none of the variance, and the best baseline (SVR on LIWC and NRC) has an MSE 2.0\% below that of FACET.

Baseline performance is inconsistent across domains. The MLPs on BERT and Mistral embeddings have negative $R^2$ for Agreeableness and Neuroticism, meaning they predict worse than the mean, and PsyAttention has negative $R^2$ in every domain. SVR on the 75 lexicon features is the only baseline with positive $R^2$ in all five domains.

\subsection{Ablation Study}

We conduct ablation experiments to examine the contribution of the individual components of the proposed framework. Table~\ref{tab:ablations} reports domain-level $R^2$ and MSE across the five Big Five traits. Unless otherwise specified, the full model uses both text and psycholinguistic modalities, concatenation for modality combination, attention-based facet pooling, and joint training with all three phases.

\textbf{Modality.} Using either modality alone generally reduces predictive performance relative to using both modalities. Text-only prediction provides higher $R^2$ than Psy-only prediction for O, C, E, and N, with differences of 158.3\%, 150\%, 70.0\%, and 100.0\%, respectively, relative to the corresponding Psy-only results. The MSE is also lower for text-only prediction for O, C, E, and N. For A, however, Psy-only prediction produces a slightly higher $R^2$ and lower MSE than text-only prediction. Overall, these results suggest that text provides a stronger contribution to personality prediction across most domains, while the psycholinguistic modality also captures complementary information, particularly for A.

\textbf{Modality combination.} We next compare concatenation and weighted combination for integrating the text and psycholinguistic representations. Relative to concatenation, weighted combination results in a 25.0\%, 33.3\%, and 33.3\% decrease in $R^2$ for O, C, and A, respectively, with corresponding increases in MSE of 13.0\%, 7.0\%, and 1.7\%. For E, weighted combination instead improves $R^2$ by 20.0\% and reduces MSE by 3.5\%. For N, the two approaches produce the same $R^2$, while weighted combination reduces MSE by 1.0\%. Overall, concatenation provides stronger and more consistent performance for O, C, and A, whereas weighted combination provides an advantage for E and a lower MSE for N.

\textbf{Facet pooling.} We further examine how the six facet representations are aggregated into the domain representation, using the concatenation combination for all pooling variants. Attention-based pooling yields the lowest MSE across all five personality domains. Compared with mean pooling, attention increases $R^2$ by 6.7\%, 50.0\%, 15.4\%, and 20.0\% for O, C, E, and A, respectively, while reducing MSE by 2.1\%, 5.9\%, 2.4\%, 1.6\%, and 0.5\% for O, C, E, A, and N. Compared with concatenation, attention produces corresponding $R^2$ improvements of 6.7\%, 80.0\%, 15.4\%, and 50.0\%, with MSE reductions of 2.5\%, 8.1\%, 2.5\%, 2.6\%, and 0.5\%. For N, $R^2$ remains near zero under all three pooling strategies. Overall, these results indicate that attention-based pooling provides a more effective way to aggregate the facet representations than either mean pooling or concatenation.

\textbf{Training regime and phase removal.} We next examine the effect of the training regime and of removing Phases 1 and 2. Joint training gives lower MSE than staged training in all five domains, by 45.2, 52.9, 30.5, 6.1, and 7.8 units for O, C, E, A, and N, respectively, with $R^2$ following the same ordering. The gap is largest for O and C and small for A and N. Removing Phases 1 and 2 leaves a regression MLP on the raw features (Section~\ref{sec:regime-phase-ablation}). Relative to this model, the joint model has lower MSE for O, C, E, and A, by 25.2, 20.9, 20.5, and 9.5 units, and about the same MSE for N, whereas the staged model has higher MSE than the raw-feature MLP for O, C, E, and N, by 20.0, 32.1, 10.0, and 9.5 units, and slightly lower MSE for A, by 3.5 units. The encoders, facet heads, and attention pooling therefore improve on a regression from the raw features when trained jointly, but not when trained in stages. For N, all configurations have $R^2$ close to zero, so their ordering is not informative.

\section{Discussion}

The results show that a representation built around facets can carry personality information from life narratives in all five domains. Predicting each domain through six contrastively supervised facet embeddings produced both facet-level and domain-level outputs from one model, and the model recovered signal most reliably for Openness and least for Neuroticism. The facet-level results add something that a domain score cannot provide. Separability varied within every domain, so facets are not interchangeable parts of a domain, and the framework indicates which of them are expressed in a narrative.

Pooling the facet embeddings by attention gave the lowest error in every domain. This agrees with the premise behind the design, that the facets of a domain are not equally recoverable from language, and it indicates that letting facets contribute unequally is a better assumption than fixing their weights in advance.

Joint training outperformed staged training. A plausible reason is that the regression loss can adjust the encoders and the facet heads together under joint training, whereas in staged training the representation is fixed by the first stage and cannot be refined by the regression objective. HG-PerCon found the opposite ordering under a self-supervised contrastive objective~\cite{li2024hgpercon}. Our contrastive objective is supervised by facet labels, so it already carries trait-relevant structure that joint training can build on. We did not isolate this mechanism.

The two views contributed differently across domains. Adding the psycholinguistic features to the text embedding helped some domains and not others, which is consistent with the finding that pretrained language models capture only part of the psychological information that lexicon features encode~\cite{zhang2024psyattention}. A participant-specific weighting of the two views improved PsyAttention's classification accuracy~\cite{zhang2024psyattention}, but in our regression setting it gave no consistent advantage over concatenation. The weighting adds a scoring network to each facet head, which may be demanding for a training set of this size, although we did not test this.

On the same corpus, Hussain et al.~\cite{hussain2026language} obtained a different ordering of domains, with Neuroticism and Extraversion recovered best. Their encoder is fine-tuned on the personality scores whereas ours is frozen, and they average over cross-validation folds whereas we use a single split. Either difference could contribute, and we did not test them, so the two sets of results are not directly comparable. Fine-tuning the language model inside our framework is a natural extension.

\section{Conclusion}

We presented a three-phase framework that estimates Big Five domain scores from life-narrative interviews by combining a language model's representation with psycholinguistic features, learning a contrastive embedding for each facet, and pooling the facet embeddings by attention. On 1{,}332 SPAN participants, the framework produced facet-level and domain-level predictions from one model, with domain-level $R^2$ from $0.008$ for Neuroticism to $0.328$ for Openness and facet-level AUC up to $0.740$. Joint training and attention pooling gave the lowest MSE in every domain. Future work includes validation on other datasets and populations, fine-tuning the language model within the framework, and evaluation across demographic groups. The code is available at \url{https://github.com/AI-for-Health-Data/OCEANprediction}.

\section*{Limitations}
The evaluation uses a single corpus of 1{,}332 participants aged 55 to 64 from one region, so the results should be read as within-distribution and may not transfer to other populations, ages, or kinds of text. Cross-dataset validation is the clearest next step. The personality labels are self-reports, whose measurement error limits the accuracy that any predictor can achieve. The LIWC and NRC features are dictionary-based, English-language resources that do not capture context.

All results come from a single split and a single seed. The held-out set (266 participants) is modest, and a small validation set was used for early stopping and checkpoint selection. We report no confidence intervals or significance tests and make many comparisons across five domains and several configurations without correction, so differences of a few hundredths in $R^2$ or a few units of MSE should be regarded as indicative and not as established. For Neuroticism, all configurations have $R^2$ close to zero, so their ordering is not informative, and several facet AUCs lie within about two standard errors of chance.

The language view is a single mean-pooled 4{,}096-dimensional transcript embedding from a frozen model, so token-level information and any adaptation to the personality scores are unavailable to the downstream layers, and this may contribute to the gap to fine-tuned models on the same corpus.

Binarizing facet scores at the median discards ordinal information. The ablations also have limited scope. Removing Phases 1 and 2 removes the encoders, the facet heads, and the pooling together, so it does not isolate the contribution of facet decomposition, and we did not ablate the contrastive loss while keeping the facet heads. Finally, we did not compare with fine-tuned encoders trained under our protocol, and published results on other datasets use different targets and metrics.

\section*{Ethical Considerations}

This study develops methods for estimating personality traits from narrative interviews for research purposes. The transcripts come from the St.\ Louis Personality and Aging Network (SPAN) study, whose participants provided informed consent, and the transcripts were anonymized before analysis. The transcripts were processed locally with an open-weight language model and were not sent to external services.
The work is intended to support the analysis of narrative patterns and not to replace human judgment.

The estimates produced here are population-level statistical associations and not clinical assessments. Predictive performance is limited even for the best domain, where Openness explains about a third of the variance, and it is weak for Agreeableness and Neuroticism. The models are therefore not suitable for individual-level or high-stakes decisions such as hiring, evaluation, health screening, or psychological profiling.

Language models and lexicon resources may inherit biases from the data on which they were built, and linguistic patterns associated with demographic or cultural background may influence predictions. The participants in this study form a narrow sample (ages 55 to 64, one region), and we did not evaluate performance across demographic groups. Subgroup evaluation, fairness auditing, and transparent reporting would therefore be required before any deployment. Real-world use would also need model documentation, ongoing evaluation, and monitoring for drift, and deploying such a system without informed consent risks covert assessment. We discourage its use for surveillance, manipulation, or consequential judgments about individuals.

\section*{Data Availability Statement}

The code is available at \url{https://github.com/AI-for-Health-Data/OCEANprediction}. The interview transcripts and personality scores are derived from the St.\ Louis Personality and Aging Network (SPAN) study~\cite{oltmanns2020personality}. Because the transcripts contain personal narratives from consenting participants, they cannot be shared publicly. Researchers can request access from original study authors subject to data-use approval.

\bibliographystyle{IEEEtran}
\bibliography{references}

\end{document}